\documentclass[10pt,twocolumn,letterpaper]{article}

\usepackage[pagenumbers]{cvpr} %
\usepackage{graphicx}
\usepackage{amsmath}
\usepackage{amssymb}
\usepackage{booktabs}

\usepackage{times}
\usepackage{epsfig}
\usepackage{graphicx}
\usepackage{amsmath}
\usepackage{amssymb}

\usepackage{comment}

\usepackage{subcaption} %
\usepackage{booktabs}    %
\usepackage[font=small]{caption}
\usepackage{multirow}
\usepackage{multicol}
\usepackage{arydshln}
\usepackage{capt-of}
\usepackage{algorithm}
\usepackage{listings}
\usepackage{tabularx}
\usepackage[table]{xcolor}
\usepackage{colortbl}
\usepackage{enumitem}
\usepackage{cuted}
\usepackage{adjustbox}
\usepackage{wrapfig}

\usepackage[leftcaption]{sidecap}

\definecolor{newlightblue}{RGB}{0,75,255}
\definecolor{newlightblue}{RGB}{0,75,255}
\usepackage[pagebackref,breaklinks,colorlinks, urlcolor={newlightblue}, citecolor={newlightblue}]{hyperref}

\usepackage[capitalize]{cleveref}
\crefname{section}{Sec.}{Secs.}
\Crefname{section}{Section}{Sections}
\Crefname{table}{Table}{Tables}
\crefname{table}{Tab.}{Tabs.}

\DeclareMathOperator*{\softmax}{softmax}
\DeclareMathOperator*{\warp}{warp}
\DeclareMathOperator*{\upsample}{upsample}

\DeclareMathOperator{\tr}{tr} %

\newcommand{\fig}[1]{Figure~\ref{#1}}

\newcommand{\tbl}[1]{Table~\ref{#1}}

\newcommand{\sect}[1]{Section~\ref{#1}}
\newcommand{\mypar}[1]{\vspace{-4mm}\paragraph{#1}}
\newcommand{\vspacefig}{\vspace{-4mm}}

\newcommand{\lphotocrw}[0]{\mathcal{L}_{\mathrm{feat}}}
\newcommand{\lsmooth}[0]{\mathcal{L}_{\mathrm{smooth}}}
\newcommand{\lbound}[0]{\mathcal{L}_{\mathrm{bound}}}

\newcommand{\lcrw}[0]{\mathcal{L}_{\mathrm{msCRW}}}

\newcommand{\supparxiv}[2]{#2}

\newcommand{\projecturl}{https://jasonbian97.github.io/flowwalk}

\usepackage{etoolbox}
\makeatletter
\AfterEndEnvironment{algorithm}{\let\@algcomment\relax}
\AtEndEnvironment{algorithm}{\kern2pt\hrule\relax\vskip3pt\@algcomment}
\let\@algcomment\relax
\newcommand\algcomment[1]{\def\@algcomment{\footnotesize#1}}
\renewcommand\fs@ruled{\def\@fs@cfont{\bfseries}\let\@fs@capt\floatc@ruled
  \def\@fs@pre{\hrule height.8pt depth0pt \kern2pt}%
  \def\@fs@post{}%
  \def\@fs@mid{\kern2pt\hrule\kern2pt}%
  \let\@fs@iftopcapt\iftrue}
\makeatother
\def\cvprPaperID{713} %
\def\confName{CVPR}
\def\confYear{2022}

\definecolor{lightblue}{RGB}{0,75,255}

\newcommand{\mysection}[1]{\vspace{-2mm}\section{#1}\vspace{-1.75mm}}
\newcommand{\mysubsection}[1]{\vspace{-1.5mm}\subsection{#1}\vspace{-1.75mm}}
\newcommand{\mysubsubsection}[1]{\vspace{-2mm}\subsubsection{#1}\vspace{-1.75mm}}

\newcommand{\sz}[0]{\footnotesize}

\newcommand{\andrew}[1]{\textcolor{blue}{}}
\newcommand{\zhangxing}[1]{\textcolor{red}{}}
\newcommand{\allan}[1]{\textcolor{green}{}}
\newcommand{\alyosha}[1]{\textcolor{red}{}}

\newcommand{\x}{X}
\newcommand{\flow}{\mathbf{f}}
\newcommand{\A}{A}

\definecolor{lightgray}{RGB}{240,240,240}

\definecolor{lightyellow}{RGB}{255,255,220}
\definecolor{lightblue}{RGB}{238,238,255}

\newcommand{\bestcell}{\cellcolor{lightgray}}
\newcommand{\secondbestcell}{}

\usepackage{mathtools}

\DeclarePairedDelimiter\abs{\lvert}{\rvert}%
\DeclarePairedDelimiter\norm{\lVert}{\rVert}%

\newcolumntype{L}[1]{>{\raggedright\let\newline\\\arraybackslash\hspace{0pt}}m{#1}}
\newcolumntype{C}[1]{>{\centering\let\newline\\\arraybackslash\hspace{0pt}}m{#1}}
\newcolumntype{R}[1]{>{\raggedleft\let\newline\\\arraybackslash\hspace{0pt}}m{#1}}

\title{Learning Pixel Trajectories with Multiscale Contrastive Random Walks}

\author{\hspace{11.1mm}Zhangxing Bian\textsuperscript{1,3}
\and Allan Jabri\textsuperscript{2}
\and Alexei A. Efros\textsuperscript{2}
\and Andrew Owens\textsuperscript{1}
\and
\\ \vspace{-7.1mm} \and University of Michigan\textsuperscript{1} %
\and UC Berkeley\textsuperscript{2}
\and Johns Hopkins University\textsuperscript{3}}

\begin{document}

\maketitle

\begin{abstract}
A range of video modeling tasks, from optical flow to multiple object tracking, share the same fundamental challenge: establishing space-time correspondence.  Yet, approaches that dominate each space differ.
We take a step towards bridging this gap by extending the recent contrastive random walk formulation to much denser, pixel-level space-time graphs. The main contribution is introducing hierarchy into the search problem by computing the transition matrix between two frames in a coarse-to-fine manner, forming a \textit{multiscale} contrastive random walk when extended in time. This establishes a unified technique for self-supervised learning of optical flow, keypoint tracking, and video object segmentation. 
Experiments demonstrate that, for each of these tasks, the unified model achieves performance competitive with strong self-supervised approaches specific to that task.\footnote{Project webpage: \url{\projecturl}}

\end{abstract}

\vspace{-2mm}
\mysection{Introduction}

Temporal correspondence underlies a range of video understanding tasks, from optical flow to object tracking. At the core, the challenge is to estimate the motion of some entity as it persists in the world, by searching in space and time. 
For historical reasons, the practicalities differ substantially across tasks: optical flow aims for dense correspondences but only between neighboring pairs of frames, whereas tracking cares about longer-range correspondences but is spatially sparse.
We argue that the time might be right to try and re-unify these different takes on temporal correspondence.  \looseness=-1

An emerging line of work in self-supervised learning has shown that generic representations pretrained on unlabeled images and video can lead to strong performance across a range of tracking tasks~\cite{he2019moco, wang2019learning,jabri2020spacetime,lai20mast, wang2021unitrack}. 
The key idea is that if tracking can be formulated as label propagation~\cite{Zhu02learningfrom} on a space-time graph, all that is needed is a good measure of similarity between nodes. %
Indeed, the recent \textit{contrastive random walk} (CRW) formulation~\cite{jabri2020spacetime} shows how such a similarity measure can be learned for temporal correspondence problems,
suggesting a path towards a unified solution.
However, scaling this perspective to 
pixel-level space-time graphs holds challenges.
Since computing similarity between frames is quadratic in the number of nodes, estimating dense motion is prohibitively expensive. Moreover, 
there is no way of explicitly estimating the motion in ambiguous cases, like occlusion.
In parallel, the unsupervised optical flow community has adopted highly effective methods for dense matching~\cite{jason2016back}, which use {\em multiscale} representations~\cite{burt1983, lucas1981iterative, Bouguet00pyramidalimplementation, sun2010secrets} to reduce the large search space,
and smoothness priors to deal with ambiguity and occlusion. 
But, in contrast to the self-supervised tracking methods, they rely on {hand-crafted} distance functions, such as the Census Transform~\cite{jonschkowski2020matters,meister2017unflow}.  %
Furthermore, because they focus on producing point estimates of motion, they may be less robust under long-term dynamics.

In this work, we take a step toward bridging the gap between tracking and optical flow by extending the contrastive random walk formulation~\cite{jabri2020spacetime} to much denser, pixel-level space-time graphs. The main contribution is introducing hierarchy into the search problem, \ie, the \textit{multiscale} contrastive random walk. By integrating local attention in a coarse-to-fine manner, the model can efficiently consider a distribution over pixel-level trajectories. Through experiments across optical flow and video label propagation benchmarks, we show:
\begin{figure*}
 \center{
 \includegraphics[width=0.85\linewidth]{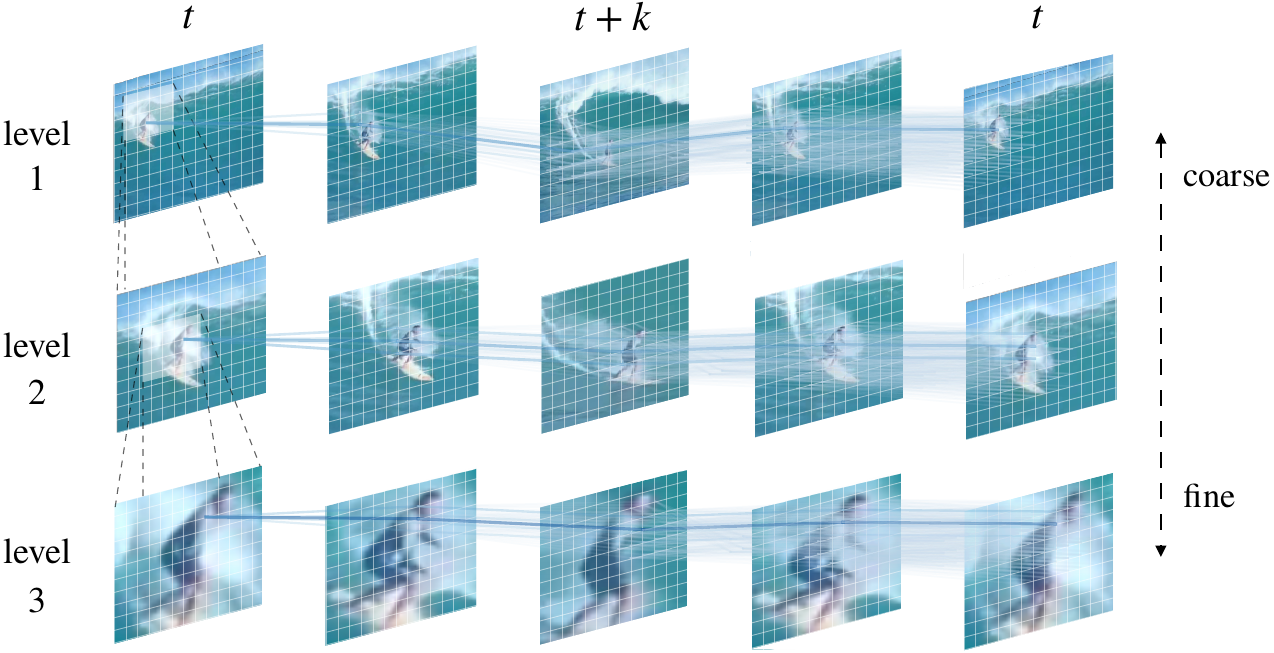}

\captionof{figure}{{\bf Multiscale contrastive random walks}. We learn representations for dense, fine-grained matching using {\em multiscale contrastive random walks}. At each scale, we create a space-time graph in which each pixel is a node, and where nodes close in space-time are connected. Transition probabilities are based on similarity in a learned representation. We train the network to maximize the probability of obtaining a cycle-consistent random walk. Here, we illustrate a pixel's walk over 3 spatial scales and 3 frames. The walker is initialized using its position at the previous, coarser scale. }} %
\label{fig:teaser}
\vspace{-4mm}
\end{figure*}

\vspace{2mm}
\begin{itemize}[leftmargin=*,topsep=1pt, noitemsep]

\item This provides a unified technique for self-supervised optical flow, pose tracking, and video object segmentation.   
\item  %
For optical flow, the model is competitive with many recent unsupervised optical flow methods, despite using a novel loss function (without hand-crafted features). %

\item For tracking, the model outperforms existing self-supervised approaches on 
pose tracking, and is competitive on video object segmentation.

\item Contrastive cycle-consistency provides a complementary learning signal to photo-consistency.
\item Multi-frame training improves two-frame performance.

\end{itemize}

\vspace{2mm}
\mysection{Related work}
\paragraph{Space-time representation learning.}
Recent work has proposed methods for tracking objects in video through self-supervised representation learning.
Vondrick~\etal~\cite{vondrick2018tracking} posed tracking as a colorization problem, by training a network to match pixels in grayscale images that have the same held-out colors. The assumption that matching pixels have the same color may break down over long time horizons, and the method is limited to grayscale inputs. This approach was extended to obtain higher-accuracy matches with two-stage matching~\cite{lai20mast,li2019joint}. In contrast to our approach, the predictions are coarse, patch-level associations.
Another line of work learns representations by maximizing cycle consistency. These methods track patches forward, then backward, in time and test whether they end up where they began.  Wang~\etal~\cite{wang2019unsupervised} proposed a method based on hard attention and spatial transformers~\cite{jaderberg2015spatial}. Jabri~\etal~\cite{jabri2020spacetime} posed cycle-consistency as a random walk problem, allowing the model to obtain dense supervision from multi-frame video. Tang~\etal~\cite{tang2021breaking} proposed an extension that allowed for fully convolutional training. %
These approaches are trained on sparse patches and learn coarse-grained correspondences. In contrast, we learn pixel-to-pixel correspondences. %
Other work has encouraged patches in the same position in neighboring frames to be close in an embedding space~\cite{gordon2020watching,xu2021rethinking}.

\mypar{Optimization-based optical flow.}

Lucas and Kanade~\cite{lucas1981iterative,baker2004lucaskanade} used Gauss-Newton optimization to minimize a brightness constancy objective. In their seminal work, Horn and Schunck~\cite{hornschunck1981hs} combined brightness constancy with a spatial smoothness criteria, and estimated flow using variational methods. Later research improved flow estimation using robust penalties~\cite{black1996robust,brox2004high,sun2008learning,sun2010secrets,liu2009beyond}, coarse-to-fine estimation~\cite{bruhn2005lucas}, discrete optimization~\cite{felzenszwalb2004efficient,shekhovtsov2008efficient,roth2009discrete}, feature matching~\cite{brox2009large,weinzaepfel2013deepflow,revaud2015epicflow}, bilateral filtering~\cite{anderson2016jump}, and segmentation~\cite{zitnick2005consistent}. In contrast, our model estimates flow using neural networks.

\mypar{Unsupervised optical flow.} Early work~\cite{memisevic2007unsupervised} used Boltzmann machines to learn transformations between images. More recently, Yu~\etal\cite{jason2016back} train a neural network to minimize a loss very similar to that of optimization-based approaches.  Later work extended this approach by adding edge-aware smoothing~\cite{wang2018occlusion}, hand-crafted features~\cite{meister2018unflow}, occlusion handling~\cite{janai2018unsupervised,wang2018occlusion}, learned upsampling~\cite{luo2021upflow}, and depth and camera pose~\cite{zhong2019unsupervised,yin2018geonet}. Another approach learns flow by matching augmented image pairs %
~\cite{liu2019ddflow,liu2019selflow,liu2020learning}. Recently, Jonschkowski~\etal~\cite{jonschkowski2020matters} surveyed previous literature and conducted an exhaustive search to find the best combination of methods and hyperparameters. %
In contrast to these works, our goal is to learn self-supervised representations for matching, in lieu of hand-crafted features. Moreover, we aim to produce a distribution over motion trajectories suitable for label transfer, rather than motion estimates alone. %
In very recent work, Stone~\etal~\cite{stone2021smurf} adapted unsupervised flow methods to the RAFT architecture~\cite{teed2020raft} and proposed new augmentation, self-distillation, and multi-frame occlusion inpainting methods. By contrast, we use PWC-net~\cite{sun2017pwcnet}, since it is the standard architecture considered in prior work, and since it can obtain strong performance with careful training~\cite{sun2021autoflow}.

\mypar{Cycle consistency.}
Cycle consistency has long been used to detect occlusions~\cite{xu2008segmentation,lei2009optical,baker2011database,hur2017mirrorflow,sundaram2010dense}, and is used to discount the loss of occluded pixels in unsupervised flow~\cite{janai2018unsupervised,wang2018occlusion,jonschkowski2020matters,ilg2018occlusions}. Zou~\etal~\cite{zou2018dfnet} used a cycle consistency loss as part of a system that jointly estimated depth, pose, and flow.
Recently, Huang~\etal~\cite{huang2021life} combined cycle-consistency with epipolar matching, but their method is weakly supervised with camera pose and assumes egomotion. 
In contrast, ours is entirely unsupervised and is capable of working {\em solely} with cycle-consistency and smoothness losses. Without extra constraints, their cycle consistency formulations have trivial solutions (\eg, all-zero flow). Other recent work~\cite{truong2021warp} learns to match images by ensuring that both an image and a warped variation of it match consistently with a second image, and Li~\etal~\cite{li2021self} used random walks with fixed transition matrices to smooth scene flow on point clouds.  A random walk formulation of cycle consistency also been used in semi-supervised learning~\cite{haeusser2017learning}, using labels to test  consistency. 

\mypar{Multi-frame matching.} %
Many methods use a third frame to obtain more local evidence for matching. Classic methods assume approximately constant velocity~\cite{sun2010layered,volz2011modeling,janai2017slow,wang2008estimating} or  acceleration~\cite{wang2008estimating,kennedy2015optical} and measure the photo-consistency over the full set of frames. %
Recently, Janai~\etal~\cite{janai2018unsupervised} proposed an unsupervised multi-frame flow method that used a photometric loss with a low-acceleration assumption and explicit occlusion handling. In contrast to these approaches, our method can be deployed using two frames at test time. We use subsequent frames as a {\em training signal}. There have also been a variety of approaches that track over long time horizons, often using sparse (or semi-dense) keypoints \cite{sand2008particle,rubinstein2012towards}. Other work chains together optical flow~\cite{wang2013dense,brox2010object,sundaram2010dense}, typically after removing low-texture regions. In contrast, our method also learns ``soft" per-pixel tracks, which convey the probability that pairs of pixels match.

\mypar{Supervised optical flow.}
Early work learned optical flow with probabilistic models, such as graphical models~\cite{freeman2000learning}. Other work learns parameters for smoothness and brightness constancy~\cite{sun2008learning} or robust penalties~\cite{li2008learning,black1996robust}. More recent methods has used neural networks. Fischer~\etal~\cite{dosovitskiy2015flownet,ilg2017flownet} proposed architectures with a built-in correlation layers. Sun~\etal~\cite{sun2017pwcnet} introduced a network with built-in coarse-to-fine matching. Recent work~\cite{teed2020raft} iteratively updates a flow with multiscale features, in lieu of coarse-to-fine matching.

\mysection{Method}

We first show how to learn dense space-time correspondences using mutiscale contrastive random walks, resulting in a model that obtains high quality motion estimates via simple nonparametric matching. We then describe how the learned representation can be combined with regression to handle occlusions and ambiguity, for improved optical flow.

\subsection{Multiscale contrastive random walks}
\label{sec:mscrw}

We review the single-scale contrastive random walk, then extend the approach to multiscale estimation.

\subsubsection{Preliminaries: Contrastive random walks} We build on the contrastive random walk (CRW) formulation of Jabri~\etal~\cite{jabri2020spacetime}. Given an input video with $k$ frames, we extract $n$ patches from each frame and assign each an embedding using a learned encoder $\phi$. 
These patches form the vertices of a graph that connects all patches in temporally adjacent frames. A random walker steps through the graph, forward in time from frames $1, 2, ..., k$, then backward in time from $k-1, k-2, ..., 1$. The transition probabilities are determined by the similarity of the learned representations:
\vspace{-4mm}
\begin{equation}
  \A_{s,t} = \softmax({\x_s \x_t^\top/\tau}),
  \vspace{-2mm}
\end{equation}
for a pair of frames $s$ and $t$, where $\x_i \in \mathbb{R}^{n \times d} $ is the matrix of $d$-dimensional embedding vectors, $\tau$ is a small constant, and the softmax is performed along each row. 
We train the model to maximize the likelihood of {\em cycle consistency}, \ie, the event that the walker returns to the node it started from:
\vspace{-1mm}
\begin{align}
\small
\mathcal{L}_\mathtt{CRW} = -\frac{1}{n}\tr(\log(\bar{A}_{t, t+k}\bar{A}_{t+k, t})),
\vspace{-2mm}
\label{eq:crw}
\end{align}
where the $\log$ is elementwise and $\bar{A}_{t, t+k}$ are the transition probabilities from frame $t$ to $t+k$: \  $\bar{A}_{t, t+k} = \prod_{i=t}^{t+k-1} A_{i,i+1}$.

\vspace{-1mm}
\subsubsection{Optical flow as a random walk}
After training, the transition matrix contains the probability that a
pair of patches is in space-time correspondence. We can estimate the
optical flow~${\flow}_{s,t}  \in \mathbb{R}^{n \times 2}$ of a patch between frames $s$ and $t$ by taking the expected value of the change in spatial position:
\begin{equation}
\vspace{-1mm}
  g_{\mathtt{avg}}(A_{s,t}) = \mathbb{E}_{A_{s,t}} [\mathbf{f}_{s,t}] = A_{s,t} D - D,
  \label{eq:avg}
\end{equation}
where $D \in \mathbb{R}^{n \times 2}$ is the (constant) matrix
containing of pixel coordinates for each patch, and $ A_{s,t} D$ is the walker's expected position in frame $t$. %

In contrast to widely-used forward-backward cycle consistency formulations~\cite{zou2018dfnet,huang2021life}, which measure the deviation of a predicted motion from a starting point, there is no trivial solution (\eg, all-zero flow). %
This is because cycle consistency is measured in an embedding space defined solely from the visual content of image regions.%

\vspace{-2mm}
\subsubsection{Multiscale random walk}
\label{sec:multiscaleest}

As presented so far, this formulation is expensive to scale to high-resolutions because computing the transition matrix is quadratic in the number of nodes. %
We overcome this by introducing hierarchy into the search problem.
Instead of comparing all pairs, we only attend on a local neighborhood. By integrating local search across scales in a coarse-to-fine manner, the model can efficiently consider a distribution over pixel-level trajectories.

\mypar{Coarse-to-fine local attention.} 
Computing the transition matrix closely resembles cost volume estimation in optical flow~\cite{brox2004high,sun2010secrets,fischer2015flownet}. This inspires us to draw on the classic spatial pyramid commonly used for multiscale search in optical flow, by iteratively computing the dense transition matrix, from coarse to fine spatial scales $l\in[1..L]$. 

For frames $s$ and $t$, we compute feature pyramids $X_s^l \in \mathbb{R}^{h^{(l)}w^{(l)} \times d}$, 
where $h^{(l)}$ and $w^{(l)}$ are the width and height of the feature map at scale $l$. 
To match each level efficiently, we 
{\em warp} the features of the target frame $X^{l}_t$ into the coordinate frame of $X^{l}_s$ using the coarse flow from the previous level 
$f^{l}_{s,t}$; we then compute \textit{local} transition probabilities on the warped feature to account for remaining motion. Thus, we estimate the transition matrix and flow in a \textit{coarse-to-fine} manner (Fig.~\ref{fig:coarse2fine}), computing levels recurrently:
\begin{figure}[t]
\centering
    \includegraphics[width=0.86\linewidth]{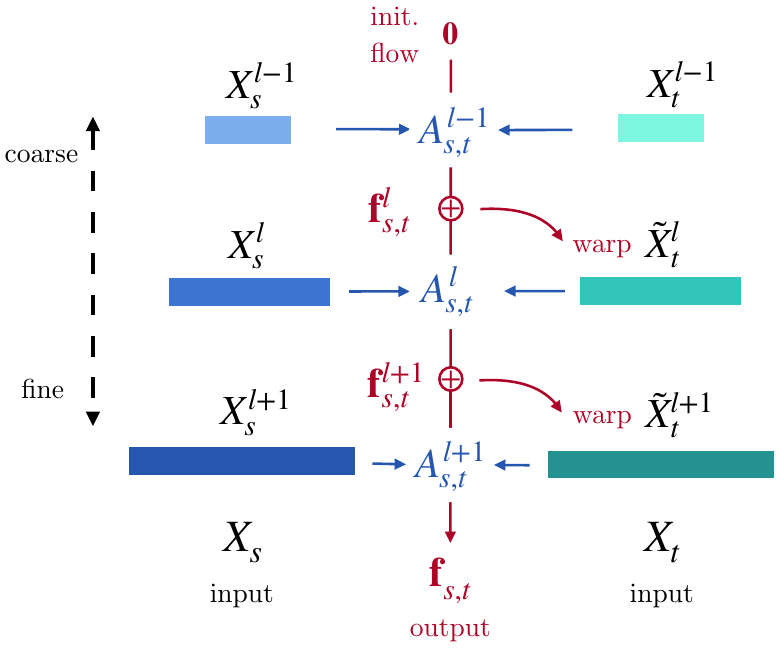}\vspace{-2mm}
    \caption{{ {\bf Coarse-to-fine matching.}} Our model performs a contrastive random walk across spatial scales: it computes a transition matrix (Eq.~\ref{eq:att}), uses it to obtain flow (Eq.~\ref{eq:upsamp}), and recurses to the next level by using upsampled estimated flow to align the finer scale for matching (Eq.~\ref{eq:warp}). \texttt{warp} samples the grid using the flow. }   %
    \label{fig:coarse2fine}
\vspacefig
\end{figure}
\vspace{-0.2em}
\begin{align}
  \tilde{X}^l_t &= \warp(X^{l}_t, f^{l}_{s,t}) \label{eq:warp}\\
    A^l_{s,t} &= \mathrm{masked\_softmax}(X^l_s \tilde{X}^{l\top}_{t} / \tau) \label{eq:att}\\
    \mathbf{f}^{l+1}_{s,t} &= \upsample(g_{\mathtt{avg}}(A^l_{s,t}) + \mathbf{f}^{l}_{s,t}) \label{eq:upsamp},
\end{align}
where $\warp(X, \mathbf{f})$ samples features $X$ with flow $\mathbf{f}$ using bilinear sampling, and $\mathbf{f}^1 = \mathbf{0}$. For notational convenience, we write the local transition constraint using $\mathrm{masked\_softmax}$, which sets values beyond a local spatial window to zero.
In practice, we use optimized correlation filtering kernels to compute Eq.~\ref{eq:att} efficiently, and represent the transition matrices $A_{s,t}^l$ as a sparse matrix. %

\mypar{Loss.} After computing the transition matrices between all pairs of adjacent frames, we sum the contrastive random walk loss over all levels:
\vspace{-2mm}
\begin{align}
    \lcrw &= - \sum_{l=1}^L \frac{1}{n_l} \tr(\log({\bar{A}}^l_{t,t+k} {\bar{A}}^l_{t+k,t})),
    \label{eq:mscrwfull}
\end{align}
where $\bar{A}_{s,t}^l$ is defined as in~Eq.~\ref{eq:crw} and $n_l$ is the number of nodes in level $l$. In our experiments, we use $L = 5$ scales and consider $k \in [2..4]$ length cycles.

\subsubsection{Smooth random walks}
Since natural motions tend to be smooth~\cite{rosenbaum2013learning}, we follow work in optical flow~\cite{sun2010secrets}, and incorporate smoothness as an additional desiderata for our random walks. We use the edge-aware loss of Jonschkowski \etal \cite{jonschkowski2020matters}, which penalizes spatial changes in flow near similarly-colored pixels:
\begin{equation}
\vspace{-1mm}
\small
\lsmooth =
\mathbb{E}_p \sum_{d \in \{x, y\}} \exp(-\lambda_c I_d(p)) \abs{ \frac{\partial^2\flow_{s,t}(p)}{\partial d^2 }}
\vspace{-1mm}
\label{eq:smooth}
\end{equation}
where $I_d(p) =  \frac{1}{3}\sum_c \abs{\frac{\partial I_c}{\partial d}}$ is the  spatial derivative averaged over all color channels $I_c$ in direction $d$. The parameter $\lambda_c$ controls the influence of similarly colored pixels. %
We apply this loss to each scale of the model.

\subsection{Handling occlusion} \label{sec:regress}
While effective for most image content, nonparametric matching has no mechanism for estimating the motion of pixels that become occluded, since it requires a corresponding patch in the next frame. We propose a variation of the model that combines the multiscale contrastive random walk with a regression module that directly predicts the flow values at each pixel.

The architecture of our regressor closely follows the refinement module of PWC-net~\cite{sun2017pwcnet}. We learn a function $g_{\mathtt{reg}}(\cdot)$ that regresses the flow at each pixel from the multiscale contrastive random walk cost-volume and convolutional features. These features are obtained from the same shared backbone that is used to compute the embeddings\supparxiv{}{ (model diagram provided in~\fig{fig:model_diagram})}. \supparxiv{We provide a diagram in the supplement.}{}

\mypar{Regression loss.} We train the regressor using a loss that closely resembles the contrastive random walk objective. Under this loss, pixels that are already well-matched by the nonparametric model (\eg, non-occluded pixels) will be unlikely to change their flow values, while poorly-matched pixels (\eg, occluded pixels) will obtain their flow values using a smoothness prior. We use the same smoothness loss, $\lsmooth$, as the nonparametric model (Eq.~\ref{eq:smooth}), and penalize the model from deviating from the nonparametric flow estimate (Eq.~\ref{eq:avg}). We also use our learned embeddings as features for a {\em learned} photometric loss, \ie, incurring loss if the model puts two pixels with dissimilar embedding vectors into correspondence. This results in an additional loss:
\begin{equation}
\small
\lphotocrw =  \norm{\x_s - \warp(\x_t, \flow_{s,t})}^2 + \lambda_a \norm{\flow_{s,t} - g_{\mathtt{avg}}(A_{s,t})}^2,
  \label{eq:photocrw}
\end{equation}
where $\flow_{s,t}$ is the predicted flow and $\lambda_a$ is a constant. To prevent the regression
loss from influencing the learned embeddings that it is based on, we do not propagate gradients
from the regressor to the embeddings $\x$ during training. As in Eq.~\ref{eq:mscrwfull}, we apply the loss to each scale and sum.

\mypar{Augmentation and masking.} We follow~\cite{jonschkowski2020matters} and improve the regressor's handling of occlusions through augmentation, and by discounting the loss of pixels that fail a consistency check~\cite{liu2019ddflow}. To handle pixels that move off-screen, we compute flow, then randomly crop the input images and compute it again, penalizing deviations between the flows. This results in a new loss, $\lbound$. We also remove the contribution of pixels in the photometric loss (Eq.~\ref{eq:photocrw}) if they have no correspondence in the backwards-in-time flow estimation from $t$ to $s$. Both are implemented exactly as in~\cite{jonschkowski2020matters}. These losses apply only to the regressor, and thus do not directly affect the nonparametric matching.

\subsection{Training}

\paragraph{Objective.}

The pure nonparametric model (\sect{sec:mscrw}) can be trained by simply minimizing the multiscale contrastive random walk loss with a smoothness penalty:
\vspace{-1mm}
\begin{equation}
  \small
    \mathcal{L}_\mathtt{non} = \lcrw + \lsmooth.
    \vspace{-2mm}
\end{equation}
Adding the regressor results in the following loss:
  \vspace{-1mm}
\begin{equation}
  \small
    \mathcal{L}_\mathtt{reg} = \lcrw + \lsmooth + \lphotocrw + \lbound.
        \vspace{-2mm}
\end{equation}
We include weighting factors to control the relative importance of
each loss \supparxiv{(specified in the supplement)}{(Section~\ref{sec:hyperparams})}.

\mypar{Architecture.} To provide a straightforward comparison with unsupervised optical flow methods, we use the PWC-net architecture~\cite{sun2017pwcnet} as our network backbone, after reducing the number of filters~\cite{jonschkowski2020matters}. This network uses the feature hierarchy from a convolutional network to provide the features at each scale. We use the cost volume features from this network as the embedding for the random walk, $\x_s^l$, after performing $\ell_2$ normalization. We also use its regressor architecture. We provide architecture details \supparxiv{in the supplementary material.}{in Section~\ref{sec:Earchitecture}.}

\mypar{Subcycles.} We follow~\cite{jabri2020spacetime} and include { subcycles} in our contrastive random walks: when training the model on $k$-frame videos, we include losses for walks of length $k$, $k-1$, ... 2. These losses can be estimated efficiently by reusing the transition matrices for the full walk.

\mypar{Multi-frame training.} When training with $k > 2$ frames, we use curriculum learning to speed up and stabilize training. We train the model to convergence with 2, 3, ... $k$ frame cycles in succession.

\mypar{Optimization.} To implement the contrastive random walk, we exploit the sparsity of our coarse-to-fine formulation, and represent the transition matrices $A_{s,t}^l$ as sparse matrices.  This significantly improved training times and reduced memory requirements, especially in the finest scales. It takes approximately 3 days to train the full model on one GTX2080 Ti. We train our network with PyTorch~\cite{paszke2019pytorch}, using the Adam~\cite{kingma2014adam} optimizer with a cyclic learning rate schedule~\cite{smith2019super} with a base learning rate of $10^{-4}$ and a max learning rate of $5 \times 10^{-4}$. \supparxiv{We provide training hyperparameters in supplementary material.}{ We provide training hyperparameters in Section~\ref{sec:hyperparams}.}

\mypar{Avoiding shortcuts.} The contrastive random walk can potentially obtain shortcut solutions when it is trained with a fully convolutional network by exploiting positional information~\cite{jabri2020spacetime}. While recent work has shown this can be solved through augmentation~\cite{tang2021breaking}, we found that we avoided trivial shortcuts when using reflection padding in our network (for all convolution layers except for in the regressor). This may be because we simultaneously optimize multiple losses and use a limited search window, making the trivial solution harder to find.

\mysection{Results}

Our model produces two outputs: the optical flow fields and the pixel trajectories (which are captured in the transition matrices). We evaluate these predictions on label transfer and motion estimation tasks. We compare them to space-time correspondence learning methods, and with unsupervised optical flow methods.

\mysubsection{Datasets}
For simple comparison with other methods, we train on standard optical flow datasets. We note that the training protocols used by unsupervised optical flow literature are not standardized. We therefore follow the evaluation setup  of~\cite{jonschkowski2020matters}. We pretrain models on unlabeled videos from  the {\bf Flying Chairs} dataset~\cite{dosovitskiy2015flying}. We then train on the {\bf KITTI-2015}~\cite{geiger2013ijrr} multi-view extension and {\bf Sintel}~\cite{butler2012sintel}. To evaluate our model's ability to learn from internet video, we also trained the model on {\bf YouTube-VOS}~\cite{xu2018youtube}, without pretraining on any other datasets.

We also evaluate our model on standard label transfer tasks. The {\bf JHMDB} benchmark~\cite{jhuang2013towards} transfers 15 body parts to future frames over long time horizons. The {\bf DAVIS} benchmark~\cite{pont-tuset2017}  transfers object masks.

\vspace{-1mm}
\subsection{Label propagation}

We evaluate our learned model's ability to perform video label propagation, a task widely studied in video representation learning work. The goal is to propagate a label map provided in an initial video frame, which might describe keypoints or object segments, to the rest of the video. 

\begin{figure*}
    \centering
    \includegraphics[width=1.0\linewidth]{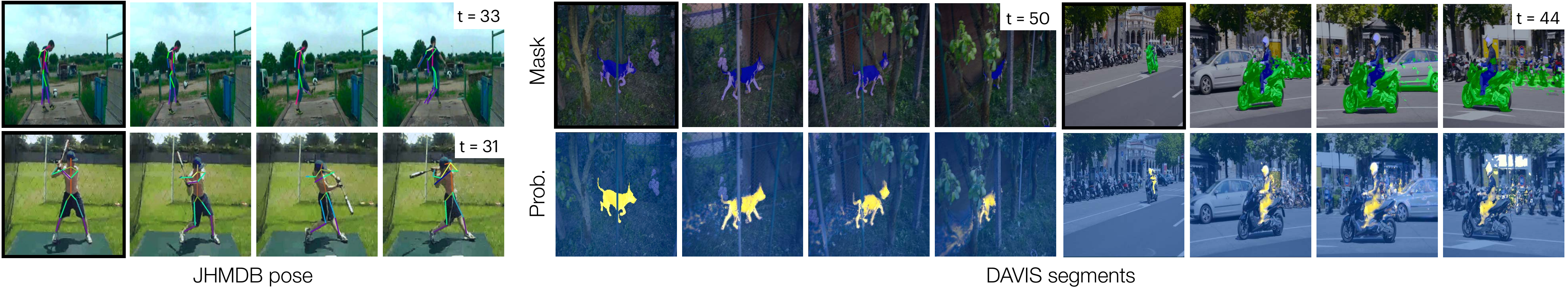}
    \vspace{-7mm}
    \caption{{\bf Propagating segments and pose along motion trajectories.} We show qualitative results for JHMDB pose (left) and object masks on DAVIS (right). For DAVIS scenes, we show examples of mask propagations (top) and {\em soft} propagated label distributions (bottom). }\vspace{-3mm}
    \label{fig:labelprop}
    \vspace{-0.3em}
\end{figure*}

We follow Jabri~\etal~\cite{jabri2020spacetime} and use our model's probabilistic motion trajectories to guide label propagation. We infer the labels for each {\em target} frame $t$ auto-regressively. For each previous {\em source} frame $s$, we have a predicted label map $L_s \in \mathbb{R}^{n \times c}$, where $n$ is the number of pixels and $c$ is the number classes. As in~\cite{jabri2020spacetime}, we compute  $K_{t,s}^l$, the matrix of weights for attention on the source frames, by keeping the top-$k$ logits in each row of $\bar{A}_{t,s}^l$. We use $K_{t,s}$ as an attention matrix for each label, \ie, $L_t = K_{t,s} L_s$. Using several source frames as context allows for overcoming occlusion.

We use variations of our model that was trained on the unlabeled Sintel and YouTube-VOS datasets, and use the transition matrix and flow fields at the penultimate level of the pyramid, \ie, level $4$. Since the transition matrix describes residual motion, we warp (\ie, with $f^4_{t, s}$) each label map before querying. Finally, since the features in level $4$ have only $16$ channels, we stack features from levels $3$ and $4$ to obtain hypercolumns~\cite{hariharan2014}, before computing attention.

\begin{table}[]
    \centering
\centering
 \small
\addtolength{\tabcolsep}{-5pt}
\caption{\textbf{Segment and Pose Propagation} with the JHMDB and DAVIS  benchmarks, respectively. $\dagger$: chained flow baseline.} \vspace{-2mm}
\begin{tabularx}{\linewidth}{l @{\extracolsep{\fill}} C{0.15\linewidth} C{0.12\linewidth} C{0.1\linewidth} C{0.1\linewidth} C{0.18\linewidth}}
\toprule
 &    & \multicolumn{3}{c}{Pose (PCK)} & Segments \\%
\cmidrule(lr){3-5}
 Method &  Arch.  & \small{@0.05} & \small{@0.1} & \small{@0.2}  & J\&F$_m$  \\ \midrule
UVC \cite{li2019joint} & ResNet18 & -- & 58.6 &  79.6 & 59.5\\ 
CRW ~\cite{jabri2020spacetime} & ResNet18 & 29.1 & {{59.3}} & {{80.6}} & {67.6}  \\ 
VFS~\cite{xu2021rethinking} & ResNet50 & -- & 60.9 & 80.7 & \bestcell {68.8} \\
\cdashline{1-6}\noalign{\vskip 1ex}
UFlow~\cite{jonschkowski2020matters}$^\dagger$ & PWC-Net & 24.1 & 51.3 & 72.1 & 42.0 \\
RAFT~\cite{teed2020raft}$^\dagger$ & RAFT & 30.2 & 55.6 & 76.0 & 46.1 \\
\cdashline{1-6}\noalign{\vskip 1ex}
{Ours - Sintel} & PWC-Net & {38.0} & \bestcell {63.1} & \bestcell {81.4} & 57.1\\
{Ours - VOS} & PWC-Net & \bestcell {38.2} & 62.6 & 80.9 & 57.9 \\
\bottomrule
\end{tabularx}

\vspace{-2em}
\label{table:parts}
\end{table}

\mypar{Evaluation.} 
We compared our model to recent video representation learning work, including single-scale CRW~\cite{jabri2020spacetime}, UVC~\cite{li2019joint}, and the state-of-the-art VFS~\cite{xu2021rethinking} (Tab.~\ref{table:parts}). We also report  two baselines that chain optical flow:  unsupervised UFlow~\cite{jonschkowski2020matters} and supervised RAFT~\cite{teed2020raft}. %

For pose propagation, we evaluate our model on JHMDB~\cite{jhuang2013towards} and report the PCK metric, which measures the fraction of keypoints within various distances to the ground truth. Our approach outperforms existing self-supervised approaches on this benchmark, especially at the stringent PCK@$0.05$, which is typically not reported. Note that our approach uses a significantly smaller network. While our model improves on fine-grained matching, it still struggles with occlusions (like other methods), which tend to involve large motions and motion blur (see Fig.~\ref{fig:labelprop}~left). For object propagation, we evaluate our model on the DAVIS~\cite{pont-tuset2017} benchmark, and report the mean of $\mathcal{J}$ and $\mathcal{F}$ metrics~\cite{perazzi2016}, which characterize segment overlap and boundary precision, respectively. Despite our focus on scaling to fine-grained matching, the model achieves competitive performance on DAVIS, outperforming the cycle-consistency method TimeCycle~\cite{wang2019learning}. In attention visualizations of the propagated label distribution, we see that the transition distribution is robust to momentary occlusion (Fig.~\ref{fig:labelprop}~mid bottom), but can nevertheless suffer from drift (Fig.~\ref{fig:labelprop}~right bottom). Interestingly, our model significantly outperforms the two optical flow methods, suggesting that the ``soft" motion trajectories provided by our model may convey useful information for propagation that is not captured by the flow.

\vspace{-1.5mm}
\subsection{Optical flow}
\vspace{-0.5mm}
We evaluate our model's ability to predict optical flow.

\begin{figure*}[t]
    \vspace{-4mm}
    \hspace{-2mm}
    \includegraphics[width=1.0\linewidth]{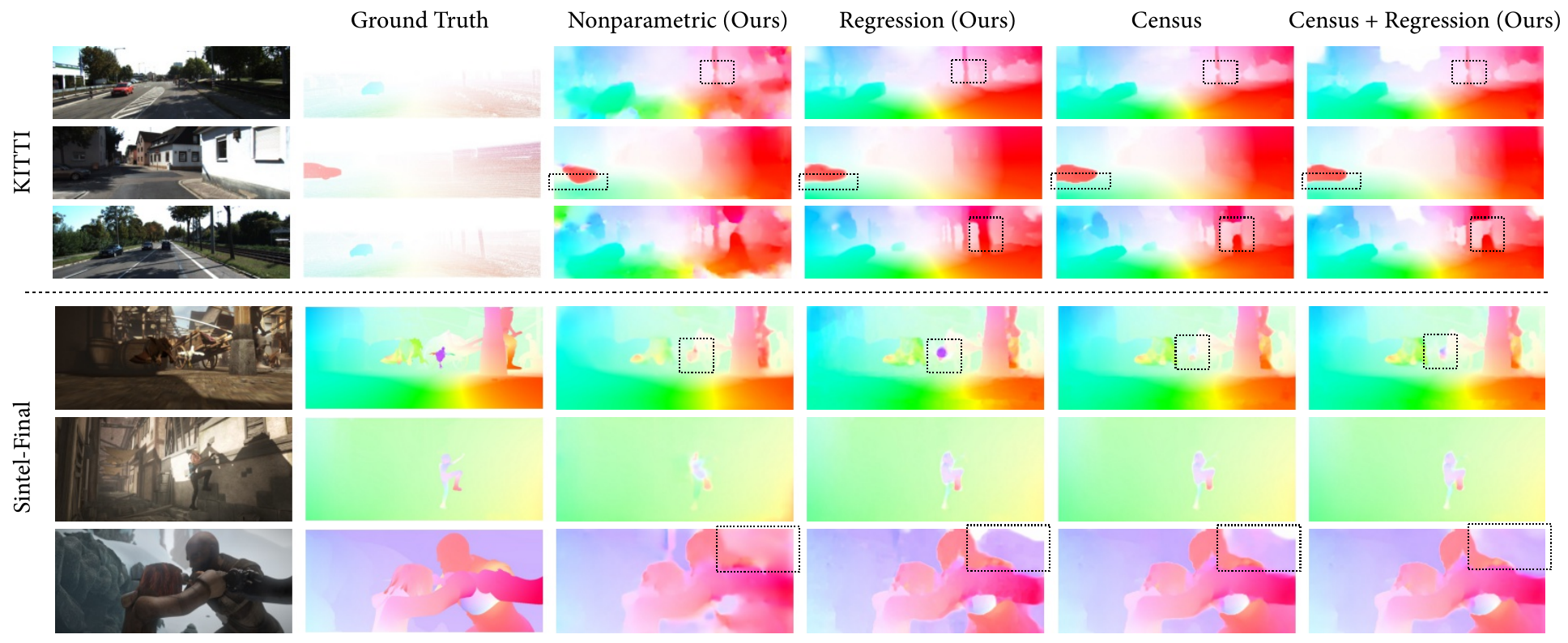}
    \vspace{-3mm}
    \caption{{\bf Optical flow qualitative results}. We show results on images not seen during training, using our nonparametric-only and regression-based models. The highlighted regions show significant differences between the regression-based models. The optical flow vectors are coded by colors.}
    \label{fig:comp}
    \vspace{-1.5em}
\end{figure*}

\vspace{-2.2mm}

\vspace{-1.5mm}
\subsubsection{Nonparametric motion estimation}
Our model is able to estimate motion solely through nonparametric matching, as can be seen qualitatively in \fig{fig:comp}. Despite the model's simplicity and the fact that it is based on very different principles than existing flow methods, it obtains strong performance on matching non-occluded pixels (Tab.~\ref{tab:CRW}). It outperforms many unsupervised optical flow models, such as SelFlow~\cite{liu2019selflow} (Tab.~\ref{tab:benchmark}) on KITTI {\em noc} metric (non-occluded endpoint error). We see that the full regression-based variation of the model obtains better results, particularly on the {\em all} metric.  

To help understand the importance of our multiscale formulation, we compared to Jabri \etal~\cite{jabri2020spacetime} on flow benchmarks, using their publicly released model (Tab.~\ref{tab:CRW}). This model resembles our nonparametric model, but with the random walk occurring at a single scale, with no smoothness prior. We evaluate the model using dense features, as in their approach. We found that our model {significantly outperforms it}. %
To control for training differences, we also tried removing scales from multiscale training (Tab.~\ref{tab:crwablations}), by using untrained (random) embeddings for the fine scales. We found that this significantly reduced performance.

\vspace{-1.5mm}
\subsubsection{Effects of multi-frame cycles}
\vspace{-1mm}
We asked how the  quality of the representation changes as we vary the number of frames $k$ used to train the random walk. We test all models on 2-frame optical flow.  As seen in Table~\ref{tab:crwablations}, the model obtains better performance on all metrics using 3-frame and 4-frame cycles. %

\vspace{-3mm}
\subsubsection{Photometric feature learning}
\vspace{-0.5em}

In contrast to other effective unsupervised optical flow approaches, our model does not use hand-crafted features. We evaluate the quality of our learned features when used as a photometric loss, compared to other common designs (Tab.~\ref{tab:photometric}). First, we compared our model to a variation of our model that uses raw pixels (rather than hand-crafted features) within a photometric loss, with a robust {\bf Charbonnier} penalty~\cite{sun2010secrets}. This baseline model is very similar to the {\em Charbonnier} variation of UFlow~\cite{jonschkowski2020matters}, though to control for other differences we use our own network. This amounts to simply replacing $\lphotocrw$ with the Charbonnier loss (which disables the contrastive random walk). 
We found that the resulting model performed significantly worse. %

Next, we considered using a state-of-the-art hand-crafted feature, the {\bf Census transform}~\cite{meister2017unflow}, resulting in a model similar to the {\em Census} variation of UFlow. We found that our features obtained competitive performance on non-occluded pixels, %
but that there was a significant advantage to Census features on the {\em all} metric. This is understandable since the contrastive random walk does not have a way of learning features for occluded pixels. Interestingly, we found that combining the two features together  improved performance, and that the gap improves further when multi-frame walks are used, obtaining the overall best results. %

Moreover, the combined features show more robustness on image pairs with rapid exposure and hue changes. We evaluated models trained on hue- and brightness-jittered image pairs and found that the model with our learned features performed significantly better, and that the gap increased with the magnitude of the jittering~(Tab. ~\ref{tab:benchmark}b). \supparxiv{Please see the supplement for details.}{Please see Section~\ref{sec:jitter} for details.}

Finally, we used our learned features as part of the photometric loss for ARFlow~\cite{liu2020learning}, a recent unsupervised flow model. We added the $\lphotocrw$ loss to their model and retrained it (while jointly learning the features through a contrastive random walk). The resulting model improves  performance on all metrics, with a larger gain on Sintel (Tab.~\ref{tab:benchmark}).

\vspace{-2mm}
\subsubsection{Motion estimation ablations}
\vspace{-1mm}

\begin{table}
	\caption{{\bf Nonparametric motion estimation}. Our single-scale, nonparametric, and regression-based methods. }
	\vspace{-2mm}
	\centering
	\adjustbox{width=0.8\linewidth}{

	\begin{tabularx}{\linewidth}{X ccc}
		\toprule
		Method  & \multicolumn{3}{c}{KITTI-15 {\em train}}  \\
		& {\sz noc} & {\sz all} & {\sz ER\% (occ)} \\ \hline
		Jabri et al.~\cite{jabri2020spacetime}      & 12.63            &     19.41     &     64.50    \\
		Ours - Nonparametric   &  2.18            &       9.42      &      27.98 \\ 
		Ours - With Regression & \bestcell 2.09& \bestcell 3.86& \bestcell 12.45  \\
		\bottomrule
	\end{tabularx}
 	}
	\vspace{-0.9em}
	\label{tab:CRW}
\end{table}

\begin{table}
	\small
	\caption{{\bf Model configurations. } All models are 2-frame. Regressor-only finds a shortcut solution, predicting zero flow.} \vspace{-2mm}
	\label{tab:core}
	\centering
	\adjustbox{width=0.9\linewidth}{
		\begin{tabularx}{\linewidth}{Xcc cc}
			\toprule
			& \phantom{a} &\multicolumn{3}{c}{KITTI-15 \textit{train}}  \\
			\cmidrule{3-5} 
			 Configuration&& noc & all & ER\%\\
			\midrule
			Full && \bestcell 2.09& \bestcell 3.86& \bestcell 12.45 \\
			Nonparametric only &&2.18&9.42&27.98 \\
			\cdashline{1-5}
			No feature consistency in $\lphotocrw$ &&2.20&5.02& 17.53\\
			Regressor only &&14.21&21.45&41.34 \\
			\cdashline{1-5}
			No regressor constraint in $\lphotocrw$ &&5.54&10.44&26.43 \\
			No $\lbound$ &&2.14&4.88&16.54 \\ 
			No $\lsmooth$ &&10.98&17.43&34.85 \\ 
			\bottomrule& & 
	\end{tabularx}
}
	\vspace{-3em}
\end{table}

 \newcommand{\mftest}[0]{$^*$}
 
\begin{table*}[ht!]
	\footnotesize

	\subfloat[][\small Optical flow benchmarks]{\resizebox{0.75\textwidth}{!}{
			\begin{tabular}{cl cccccccc}
				\toprule
				&  & \multicolumn{4}{c}{\bf Sintel}  & \multicolumn{4}{c}{\bf KITTI-15} \\
				& & \multicolumn{2}{c}{Clean} & \multicolumn{2}{c}{Final}\\
				\cmidrule(lr){3-6} \cmidrule(lr){7-10} 
				&  &  {\em train} & {\em test} & {\em train} & {\em test} & \multicolumn{3}{c}{{\em train}} & {\em test} \\
				& \textbf{Method} &  EPE & EPE & EPE & EPE & all & noc & ER \% & ER \% \\%
				\hline
				\parbox[c]{2mm}{\multirow{4}{*}{\rotatebox[origin=c]{90}{Supervised~}}} & FlowNetC \cite{dosovitskiy2015flownet}  & (3.78) & 6.85 & (5.28) & 8.51 & - & - & - & - \\
				& FlowNet2 \cite{ilg2017flownet} & \secondbestcell (1.45) & 4.16 & (2.01) & 5.74 & (2.30) & - & (8.61) & 11.48 \\
				& PWC-Net \cite{sun2017pwcnet}  & (1.70) & 3.86 & (2.21) & 5.13 & (2.16) & - & (9.80) & 9.60 \\
				& RAFT \cite{teed2020raft}  & {\bestcell (0.76)} & {\bestcell 1.94} & {\bestcell (1.22)} & {\bestcell 3.18} & {\bestcell (0.63)} & - & {\bestcell (1.50)} & {\bestcell 5.10} \\ \hline
				\parbox[t]{2mm}{\multirow{8}{*}{\rotatebox[origin=c]{90}{ Unsupervised}}}  
				& MFOccFlow\cite{Janai2018ECCV}\mftest & \{3.89\} & 7.23 & \{5.52\} & 8.81 & {[}6.59{]} & {[}3.22{]} & - & 22.94 \\
				& EPIFlow \cite{zhong2019unsupervised}  & 3.94 & 7.00 & 5.08 & 8.51 & 5.56 & 2.56 & - & 16.95 \\
				& DDFlow \cite{DDFlow}  & \{2.92\} & \secondbestcell 6.18 & \{3.98\} & 7.40 & {[}5.72{]} & {[}2.73{]} & - & 14.29 \\
				& SelFlow \cite{liu2019selflow}\mftest  & \secondbestcell {[}2.88{]} & {[}6.56{]} & \secondbestcell \{3.87\} & \secondbestcell \{6.57\} & {[}4.84{]} & {[}2.40{]} & - & 14.19 \\
				& UFlow \cite{jonschkowski2020matters}  &  \{2.50\} &  5.21 &  \{3.39\} &  6.50 &  \{2.71\} &  \{1.88\} &  \{9.05\} &  11.13 \\ 
				&SMURF-PWC\cite{stone2021smurf} &2.63&-&3.66&-&2.73&-&9.33&- \\
				&SMURF-RAFT\cite{stone2021smurf} \bestcell &\bestcell\{1.71\}&\bestcell3.15&\bestcell\{2.58\}&\bestcell4.18&\bestcell\{2.00\}&\bestcell\{1.41\}&\bestcell\{6.42\}&\bestcell6.83 \\
				& ARFlow \cite{liu2020learning}  &[2.79]&[4.78]&[3.73]&[5.89]&[2.85]&-&-& [11.80]\\
				\cdashline{1-10} 
				\parbox[t]{2mm}{\multirow{2}{*}{\rotatebox[origin=c]{90}{Ours}}} & Ours + ARFlow &  [2.71] & [4.70] & [3.61]  & [5.76] &   [2.81]   & [2.17] &[11.25]&  [11.67] \\
				& Ours (2-cycle) &\{2.84\} & 5.68 & \{3.82\} & 6.72 &   \{3.86\}  & \{2.09\} & \{12.45\} & 13.10\\
				\bottomrule
	\end{tabular}}} 
	\hspace{1mm}
	\subfloat[][\small Robustness to jittering]{\includegraphics[width=0.21\linewidth]{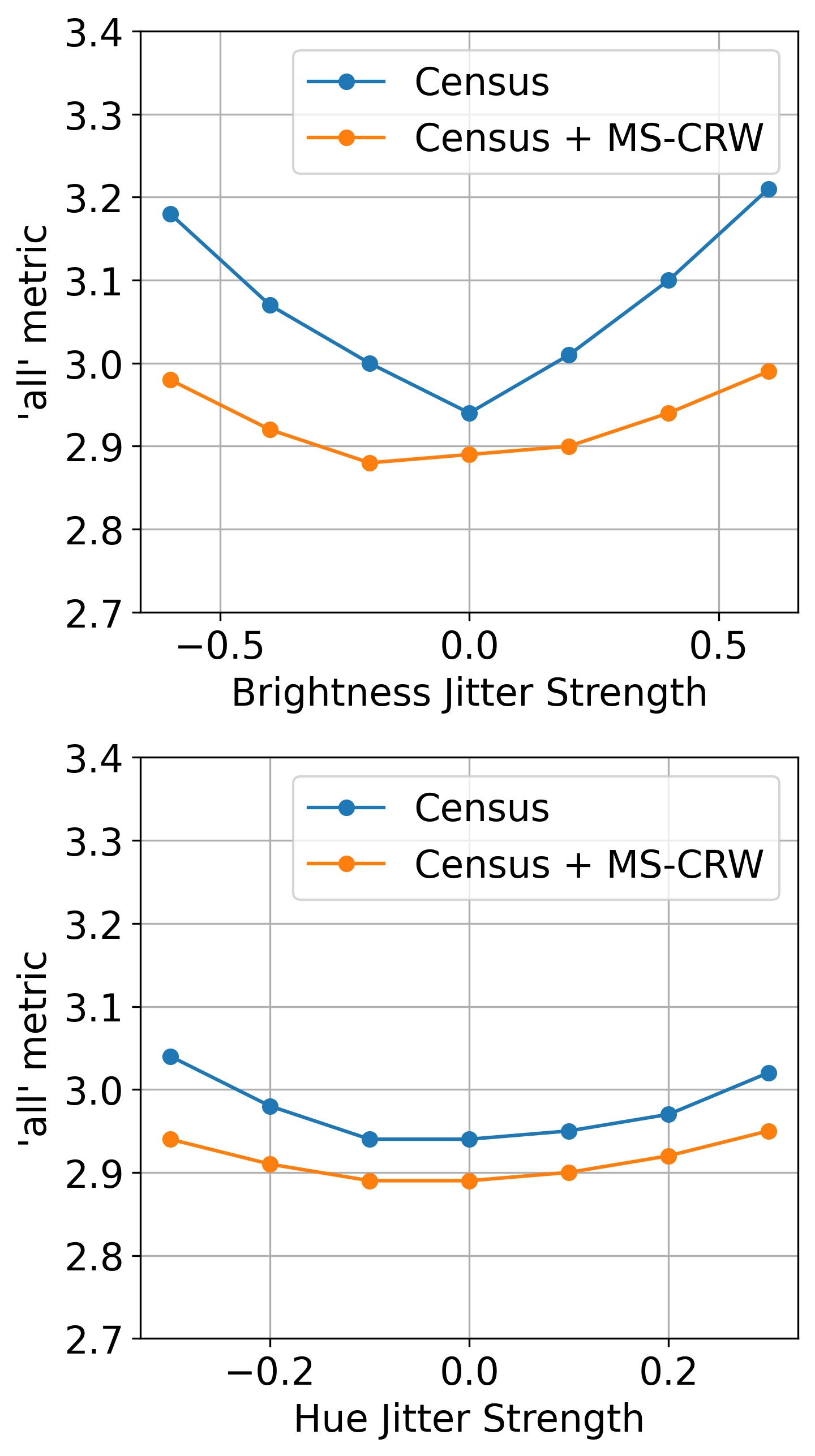}}

	\vspace{-3mm}
	\caption{(a) Our model is trained on the train/test splits of the corresponding datasets. We reprint numbers from~\cite{jonschkowski2020matters} and adopt their convention that ``\{\}'' are trained on unlabeled evaluation data, ``[]'' are trained on related data (e.g., full Sintel movie) and ``()" are supervised. Methods that use 3 frames at test time are marked with $*$. (b) We evaluate robustness to brightness and hue jittering. } 
	\label{tab:benchmark}
	\vspace{-4mm}
\end{table*}

\begin{table}
\small
\centering
	\caption{{\bf Photometric features.} We evaluate the effectiveness of our features when they are used to define a photometric loss.} \vspace{-2mm}
	\label{tab:photometric}

	\adjustbox{width=0.9\linewidth}{
		\begin{tabular}{lcc cc}
			\toprule
			& \phantom{a} &\multicolumn{3}{c}{KITTI-15 \textit{train}} \\
			\cmidrule{3-5} 
			 Losses&& noc & all & ER\%\\
			\midrule
		    Charbonnier &&2.28&5.69& 19.30 \\
			Census &&2.05&3.14&11.04 \\
			Our feats. &&2.09&3.86&12.45 \\
			\cdashline{1-5}
			Our feats. + Charbonnier &&2.21&4.51& 14.25\\
			Our feats. + Census &&2.04&3.10& 10.89\\
			Our feats. (3-frame) + Census && \bestcell 2.02 & \bestcell 3.03& \bestcell 10.67 \\
			
			\bottomrule& 
	\end{tabular}}
	\vspace{-1em}
\end{table}

\begin{table}
\footnotesize
\centering
	\caption{{\bf Contrastive random walk ablations.} We evaluate different model parameters, including cycle length and number of scales in the multiscale random walk (from coarse to fine). }\vspace{-2mm}
	\small
\footnotesize
	\adjustbox{width=0.45\linewidth}{
		\begin{tabular}{llcccc}
			\toprule
			&& & \multicolumn{3}{c}{KITTI-15 \textit{train}}\\
			& && noc & all & ER\%\\
			\midrule
				\parbox[t]{2mm}{\multirow{3}{*}{\rotatebox[origin=c]{90}{\footnotesize Cyc. len.}}} 
			&2&&2.09&3.86&12.45 \\
			&3&&2.05&3.46&12.28 \\
			&4&& \bestcell 2.04&\bestcell 3.39& \bestcell 12.14 \\

			\bottomrule& 
	\end{tabular}}
\quad
\adjustbox{width=0.45\linewidth}{
\begin{tabular}{llcccc}
	\toprule
	&& & \multicolumn{3}{c}{KITTI-15 \textit{train}}\\
	& && noc & all & ER\%\\
	\midrule
	\parbox[t]{2mm}{\multirow{3}{*}{\rotatebox[origin=c]{90}{\footnotesize \# levels}}}
	&	1 && 4.45 & 8.98&  24.52 \\
	& 3 && 2.36&  4.55&  14.35 \\
	& 5 && \bestcell 2.09& \bestcell 3.86& \bestcell 12.45  \\
	\bottomrule& 
\end{tabular}}

	\vspace{-1em}
	\label{tab:crwablations}
\end{table}

To help understand which properties of our model contribute to its performance, we perform an ablation study on  KITTI-15 (\tbl{tab:core}). We asked how the different losses contribute to the performance. We ablated the smoothness loss (Eq.~\ref{eq:smooth}), the self-supervision loss (Sec.~\ref{sec:regress}), and removing the constraint on the regressor in Eq.~\ref{eq:photocrw} by setting $\lambda_a = 0$. We see that the smoothness loss significantly improves results.  Similarly, we discarded the feature consistency term from Eq.~\ref{eq:photocrw}, which reduces the quality of the results but outperforms the nonparametric model on the {\em all} metric.

\mypar{Training on internet video.} %
We found that our model generalized well to benchmark datasets when training solely on YouTube-VOS~\cite{xu2018youtube} (Tab.~\ref{tab:vos}). For comparison, we also trained ARFlow~\cite{liu2020learning} on YouTube-VOS.  Our model obtained better performance on KITTI, while ARFlow performed better on Sintel.

	\begin{table}%
		\small
		\caption{{\bf Training on internet video.} We train on YouTube-VOS~\cite{xu2018youtube} and evaluate on optical flow benchmarks.  }\vspace{-3mm}
		\label{table:vos}
		\centering
		\adjustbox{width=0.9\linewidth}{
			\begin{tabular}{lccccccc}
				\toprule
				& \phantom{a} & \multicolumn{2}{c}{Sintel \textit{train}} & \phantom{a} & \multicolumn{3}{c}{KITTI-15 \textit{train}}\\
				&& Clean & Final && noc & all & ER\%\\
				\midrule
				Ours (2-cycle) && 3.37&4.65&& \bestcell 2.32&\bestcell 5.73& \bestcell 14.69 \\
				ARFlow~\cite{liu2020learning} && \bestcell 3.22 & \bestcell 4.51 && 2.65& 6.01& 16.47 \\
				\bottomrule 
		\end{tabular}}
		\vspace{-1.5em}
		\label{tab:vos}
	\end{table}

\mysubsubsection{Comparison to recent optical flow methods}

To help understand our model's overall performance, we compare it to recent optical flow methods (\tbl{tab:benchmark}). We include models that use different numbers of frames for the random walk, and a variation of ARFlow~\cite{liu2020learning} that uses our self-supervised features to augment its photometric loss. 

We found that our models outperform many recent unsupervised optical flow methods, including the (3-frame) {\bf MFOccFlow}~\cite{Janai2018ECCV} and {\bf EPIFlow}. In particular, we significantly outperform the recent {\bf SelFlow} method~\cite{liu2019selflow}, despite the fact that it takes 3 frames as input at {\em test} time and uses Census Transform features. By contrast, our model {\em uses no hand-crafted image features}. %
The highest performing method is the very recent, highly optimized {\bf SMURF} model\cite{stone2021smurf}, which uses the RAFT~\cite{teed2020raft} architecture instead of PWC-net and which extends UFlow~\cite{jonschkowski2020matters}. This model uses a variety of additional training signals, such as extensive data augmentation, occlusion inpainting with multi-stage training, self-distillation, and  hand-crafted features. %

\mysection{Discussion}
We have proposed a method for learning dense motion estimation using multiscale contrastive random walks. 
We see our work as a potential step toward unifying self-supervised tracking and optical flow. %
Moreover, the model can learn from internet video, which suggests that the emergent representations of such a hierarchical tracker may learn interesting part-whole structure at scale.
\mypar{Limitations and impact.} Motion analysis has many applications, such as in health monitoring, surveillance, and security. %
There is also a potential for the technology to be used for harmful purposes if weaponized. The released models are limited in scope to the datasets used in training. %

\mypar{Acknowledgements.} We thank David Fouhey and Jeff Fessler for the helpful feedback. AO thanks Rick Szeliski for introducing him to multi-frame optical flow. This research was supported in part by  Toyota Research Institute, Cisco Systems, and Berkeley Deep Drive.

{\small
\bibliographystyle{plain}
\bibliography{flow}
}

\appendix

\begin{table}[t]
	\small
	\caption{{\bf Window size ablations.} We evaluate different window sizes. }\vspace{-2mm}
	\small
	\adjustbox{width=\linewidth}{
		\begin{tabular}{llccccccc}
			\toprule
			& & \phantom{a} & \multicolumn{2}{c}{Sintel \textit{train}} & \phantom{a} & \multicolumn{3}{c}{KITTI-15 \textit{train}}\\
			\cmidrule{4-5}  \cmidrule{7-9} 
			& && Clean & Final && noc & all & ER\%\\
			\midrule
			\parbox[t]{2mm}{\multirow{3}{*}{\rotatebox[origin=c]{90}{\footnotesize Win. size}}}
			&$3 \times 3$ &&8.43&11.43&&5.18&7.21&30.85 \\
			&$7 \times 7 $&&3.02&4.13&&2.24&4.40&14.53 \\
			&$11 \times 11$ &&\bestcell2.84&\bestcell3.82&&\bestcell2.09&\bestcell3.68&\bestcell12.45 \\
			\bottomrule& 
	\end{tabular}}
	\vspace{-1em}

\end{table}
	
	\begin{table}
		\centering
		\small
	\caption{{\bf Hyperparameters.} We list the hyperparameters that we considered in our experiments. } 
		\begin{tabularx}{\linewidth}{ll}
			\toprule
			{\bf Hyperparameter} & {\bf Values}  \\
			\midrule
			Learning rate schedule & CyclicLR \\
			Base learning rate &  $10^{-4}$\\ 
			Max learning rate & $5\times10^{-4}$ \\

			Temperature $\tau$ & 0.07 \\ 
			Video length & \textbf{2, 3, 4} \\ 
			Window Size ($k\times k$)& \textbf{3,7,11} \\ 
			\toprule
			{\bf Loss weight} & {\bf Values}  \\
			\midrule
			Contrastive random walk loss~~~~ & 1 \\
			Smoothness loss & 30 \\
			Boundary loss & 1 \\
			Regressor constraint loss & 1 \\ 
			feature consistency loss & 0.1 \\
			$\lambda$ in smoothness loss & 150 \\
			\bottomrule
		\end{tabularx}
	\end{table} 

	\begin{figure}[t]
	\centering
	\includegraphics[width=\linewidth]{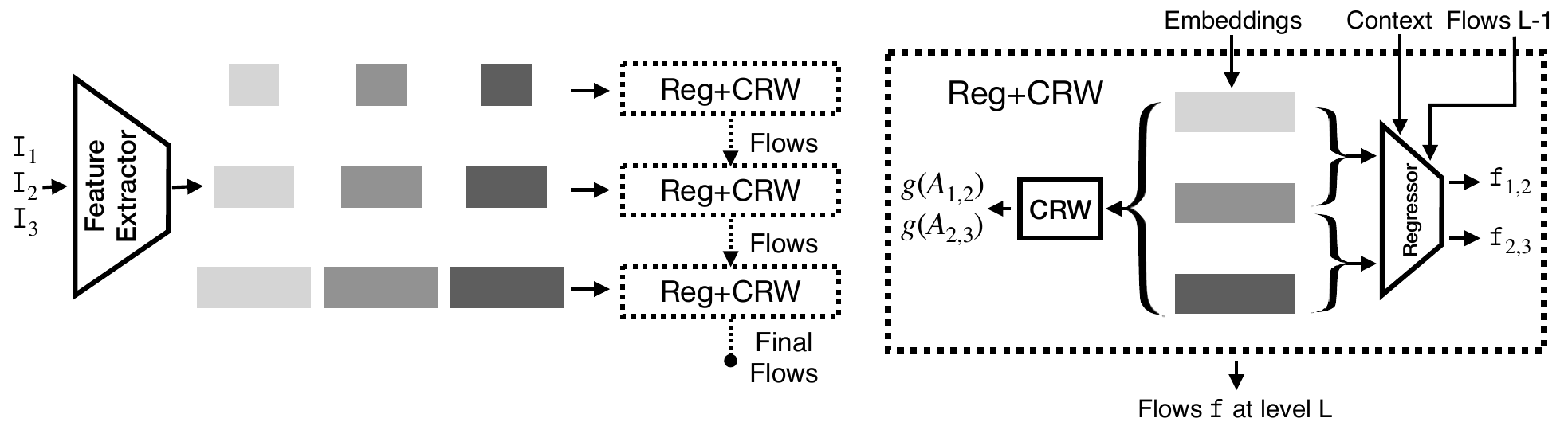} 
    \caption{{\bf Model diagram}. We extract features for the image sequence (3 images are shown here). Each spatial scale learns embeddings for CRW and flow regression. } %
	\label{fig:model_diagram}
\end{figure}

	\section{Architecture} \label{sec:Earchitecture}
	We provide additional details about the network architecture, which closely resembles PWC-Net with the simplifications introduced by ARflow~\cite{liu2020learning}. We attach a $1\times1$ convolutional layer to the each scale to obtain the embeddings (of 32 channels) for contrastive random walk at each scale. We show a diagram for the model (with the regressor) in Fig.~\ref{fig:model_diagram}.

	\begin{figure}[t]
		\centering
		\includegraphics[width=\linewidth]{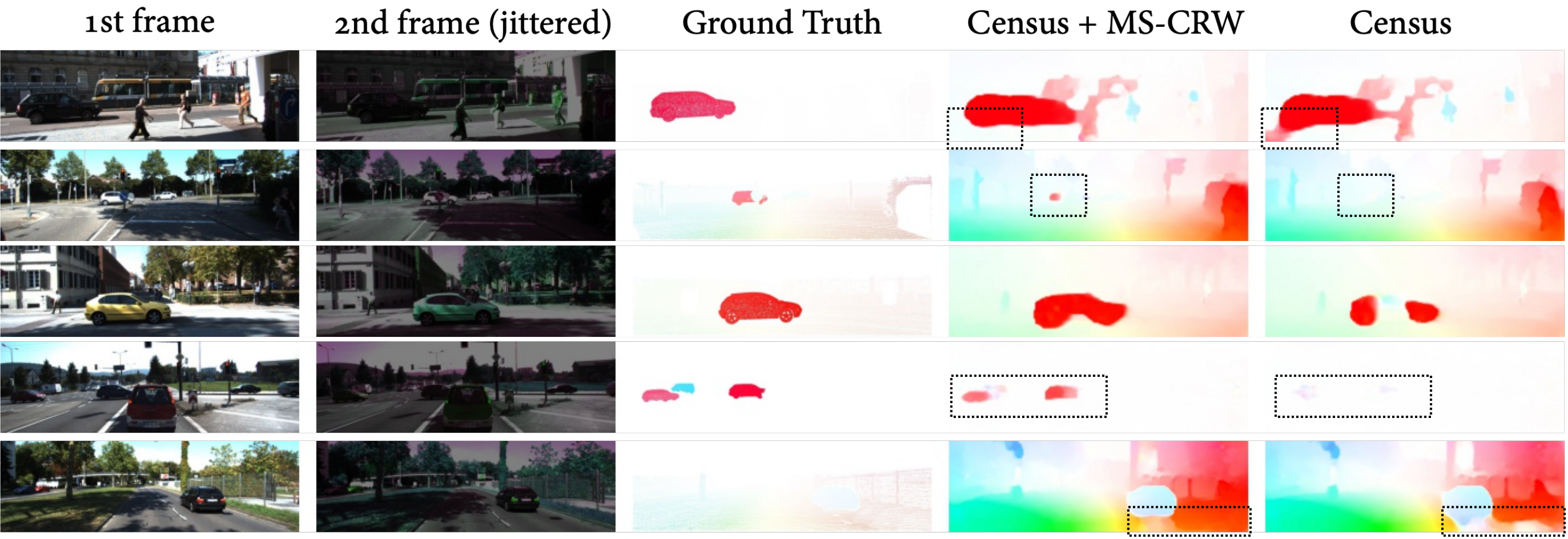} 
		\caption{\textbf{Varying brightness and hue.} Qualitative comparison between the hand-crafted Census transform feature and a model that combines these features with our self-supervised features. } 
		\label{fig:jitter}
	\end{figure}

	\paragraph{Training details.} We train our network with PyTorch~\cite{paszke2019pytorch}, using the Adam~\cite{kingma2014adam} optimizer with a cyclic learning rate schedule~\cite{smith2019super} with a base learning rate of $10^{-4}$ and a max learning rate of $5 \times 10^{-4}$.  We use batch size of 4 for 2-cycle model and 2 for 3- and 4-cycle models (due to memory constraints). The total training takes approximately four days on two GTX 2080Ti, two days for training on Flying Chairs and two days for training on Sintel/KITTI. Experiments on Sintel and KITTI start from a model that was first trained on Flying Chairs, as in~\cite{jonschkowski2020matters}.

\begin{figure*} %
 \centering
 \includegraphics[width=\linewidth]{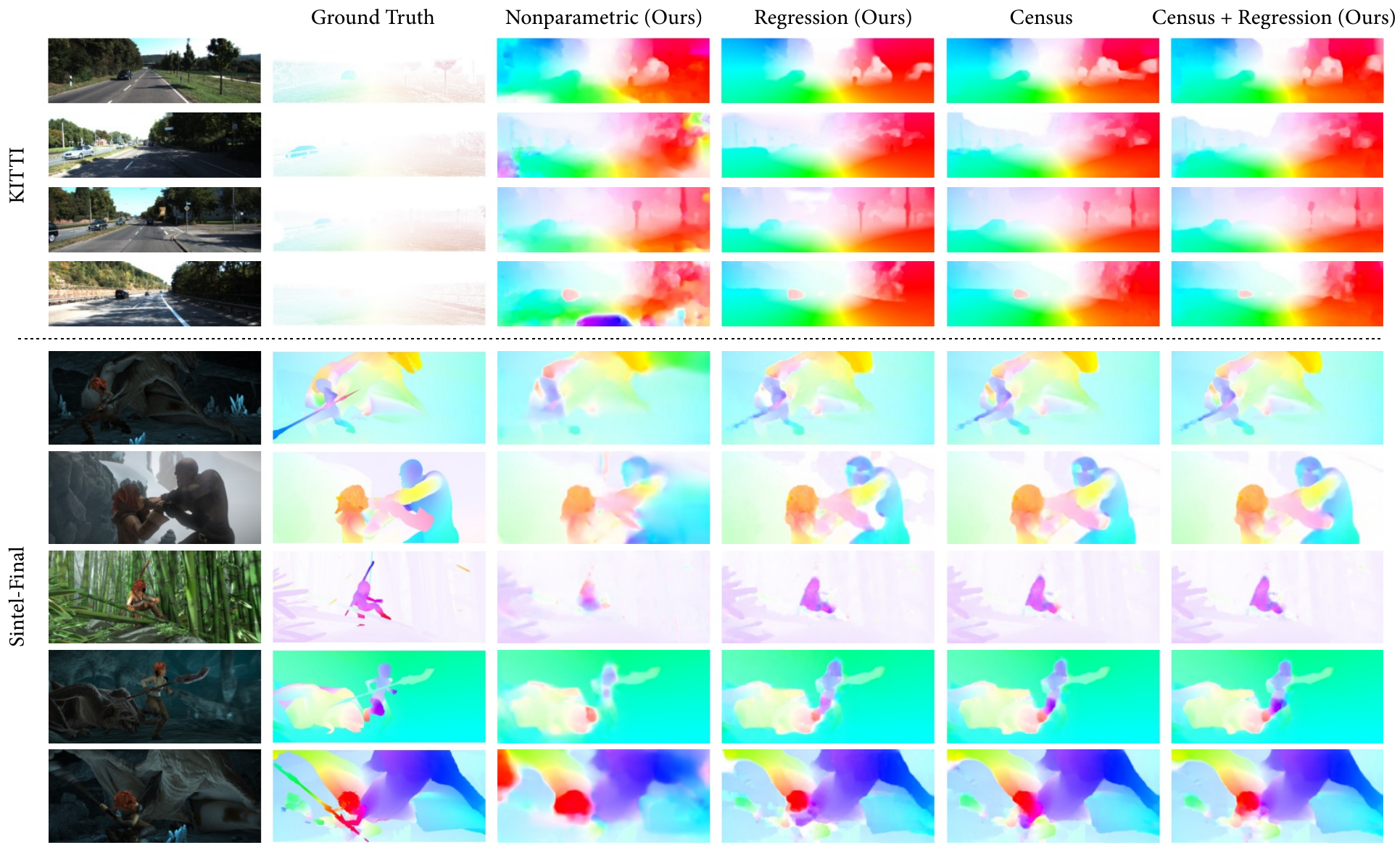}
\caption{Additional qualitative results for KITTI and Sintel optical flow.}
\label{fig:appendixflowqual}
\end{figure*}

	\section{Hyperparameters} \label{sec:hyperparams}

	We list the hyperparameters and ranges considered during our experiments. Weights for the boundary loss, learned photometric loss are hand-chosen. Parameters in bold are the ones that were systematically explored via ablations. For the image resolution, we follow the experimental setup of Jonschkowski et al.~\cite{jonschkowski2020matters} \ie, Flying Chairs: $384 \times 512$, Sintel: $448 \times 1024$, KITTI: $640 \times 640$. We use the same loss weight across different scales for a specific type of loss. RGB image values are scaled to $[-1,1]$ and augmented by randomly shifting the hue and brightness and randomly flipped left/right. The augmentation is kept the same across frame in a pair of images. In contrast to other work~\cite{jonschkowski2020matters}, we did not modify the model per dataset. For the optical flow baselines in label propagation (Table~\ref{fig:labelprop}) we used the supervised RAFT trained on FlyingThings3D~\cite{mayer2016large}.

\section{Robustness on jittered images} \label{sec:jitter} To help understand the flexibility of our self-supervised model, we trained a variation of our model on images with large, simulated brightness and hue variations, inspired by the challenges of rapid exposure changes (\eg, in HDR photography). During training and testing, we randomly jitter the brightness and hue of the second image in KITTI by a factor of up to 0.6 and 0.3 respectively, using PyTorch's~\cite{paszke2019pytorch} built-in augmentation. We finetuned the variation of our model that combines our learned features with Census features (Tab. \ref{tab:photometric}), since it obtained strong performance on KITTI. We also finetuned a model with only Census features. We found that the model with our learned features performed significantly better, and that the gap increased with the magnitude of the jittering~(Tab.~\ref{tab:benchmark}b). We show qualitative results in Fig. \ref{fig:jitter}.

    \section{Additional qualitative results}
    We provide additional qualitative results for optical flow in Figure~\ref{fig:appendixflowqual}. %

\end{document}